# Aplicación de Robots Humanoides como Guías Interactivos en Museos: Una Simulación con el Robot NAO

Sodre Hiago, hiago.sodre@estudiantes.utec.edu.uy[1]
Moraes Pablo, pablo.moraes@estudiantes.utec.edu.uy[1]
Rodríguez Monica, monica.rodriguez@utec.edu.uy[1]
Castelli Victor, victor.castelli@estudiantes.utec.edu.uy[1]
Barboza Pamela, pamela.barboza@utec.edu.uy[1]
Mattos Martin, martinmattos935@gmail.com[2]
Vivas Guillermo, greatebrte@gmail.com[2]
De vargas Bruna, bruna.devargas@utec.edu.uy[1]
Dörnbach Tobias, t.doernbach@ostfalia.de[3]
Grando Ricardo, ricardo.bedin@utec.edu.uy[1]
[1]Universidad Tecnológica del Uruguay
[2]Universidad del Trabajo del Uruguay
[3]Ostfalia University of Applied Sciences

**Abstract:** *This article presents an application that evaluates the feasibility of humanoid robots as interactive guides in art museums. The application entails programming a NAO robot and a chatbot to provide information about art pieces in a simulated museum environment. In this controlled scenario, the learning employees interact with the robot and the chatbot. The result is a skilled participation in the interactions, along with the effectiveness of the robot and chatbot that communicates the basic details of the art objects. You see natural and fluid interactions between the students and the robot. This suggests that the addition of humanoid robots to museums may provide a better experience for visitors, but also the need to continue to do more to optimize the quality of interaction. This study contributes to understanding the possibilities and requirements of applying humanoid technologies in a cultural context.*

Keywords: Educational technology, Visitor experience, Technology integration

**Resumen:** *Este artículo presenta una aplicación que evalúa la facticidad de robots humanoides como guías interactivos en varios museos de arte. La aplicación implica en programar un robot NAO V6 y un chatbot para proporcionar información sobre obras de arte en un entorno de un museo simulado. En este escenario controlado, los empleados que aprenden interactúan con el robot y el chatbot. El resultado es una participación cualificada en las interacciones, junto con la eficacia del robot y del chatbot que comunica los detalles básicos de los objetos de arte. Fueron observadas interacciones naturales y fluidas entre los estudiantes y el robot. Esto sugiere que la incorporación de robots humanoides a los museos puede proporcionar una mejor experiencia a los visitantes, pero también la necesidad de seguir haciendo más para optimizar la calidad de la interacción. Este estudio contribuye a comprender las posibilidades y requisitos de aplicar tecnologías humanoides en un contexto cultural.*

Palabras clave: Tecnología educativa, Experiencia del visitante, Integración tecnológica

## 1 - INTRODUCCIÓN

Los estudios relacionados con robots sociales se están llevando a cabo de manera creciente y su utilización en la vida cotidiana se ha convertido en una necesidad real (Hellou et al., 2022). Diversas aplicaciones pueden aprovechar la interacción humano-robot, ya que esta puede agregar un toque de innovación y modernidad a la experiencia de visitantes y usuarios. En particular, los museos podrían sacar provecho de la robótica social al crear robots para desempeñar roles de guías, educadores, animadores o incluso combinar estos servicios (Gasteiger et al., 2021). La unión de tecnología robótica y educación cultural ha originado un campo interdisciplinario centrado en la interacción humano-robot en entornos educativos. Robots humanoides como guías en espacios públicos han despertado el interés de investigadores y profesionales por su potencial educativo y cultural. Entre robots humanoides, el robot NAO, de SoftBank Robotics Europe, ofrece reconocedor y sintetizador de voz incorporados (Hellou, Mehdi et al., 2022) además de fácil programabilidad que permiten una interacción más natural con los seres humanos.

La interacción humano-robot basada en modelos de Procesamiento de Lenguaje Natural (PLN) potencia experiencias de visitantes más orgánicas mediante diversas tecnologías (Hellou et al., 2022). Estas interacciones pueden aportar autenticidad a la experiencia cultural, atraer a nuevas generaciones y realzar el valor de los centros culturales como espacios de aprendizaje y descubrimiento. En este contexto, chatbots son uno de los ejemplos más elementales y extendidos de esta interacción inteligente (Bansal y Khan, 2018). Chat GPT, desarrollado por OpenAI en 2022, ha revolucionado recientemente el mundo de los chatbots y la inteligencia artificial (Tlili, 2023). Gracias a técnicas avanzadas de PNL, Chat GPT ha demostrado una capacidad notable para generar respuestas coherentes y contextualizadas. Al aprovechar las capacidades de la robótica y el innovador modelo de Chat GPT, nuestro objetivo es verificar la viabilidad de estas tecnologías en un entorno educativo y cultural a partir del desarrollo e integración de un robot humanoide NAO V6 con Chat GPT. En este sentido exploramos su empleo en un escenario simulado donde el robot es programado para asumir el papel de guía, brindando información sobre las obras de arte en un museo hipotético. Mediante una simulación con el robot NAO, se investiga la interacción entre robots y visitantes en museos de arte, y cómo esto podría influenciar la percepción y experiencia del usuario. Esta investigación ilumina el potencial de los robots humanoides como guías en contextos educativos y aborda desafíos en la interacción humano-guía. En general, el estudio busca fundamentar futuros avances en esta área emergente.

## 2 - METODOLOGÍA

La simulación se desarrolló en un entorno real a pequeña escala con el objetivo de evaluar las posibilidades que los robots humanoides pueden aportar como guías en contextos educativos y culturales. Las subsecciones a seguir presentan el Robot NAO, su programación e integración con Chat GPT y una descripción del ambiente simulado que son desarrolladas durante todo este proyecto, tomando en cuenta análisis y resultados obtenidos en pruebas prácticas con el robot.

### 2.1 Robot NAO

Partiendo de la necesidad de trabajar con un robot completo en cuestión de funcionamiento y características de aplicación, en esta investigación fue utilizado el robot humanoide NAO V6. NAO es conocido por su apariencia antropomórfica y su capacidad para interactuar con los humanos de manera cercana, este robot es la sexta generación de este robot y se utiliza en una variedad de aplicaciones, que van desde la educación hasta la investigación y el entretenimiento. La Figura 1 presenta una imagen del robot comentado que se encuentra en la Figura 1.

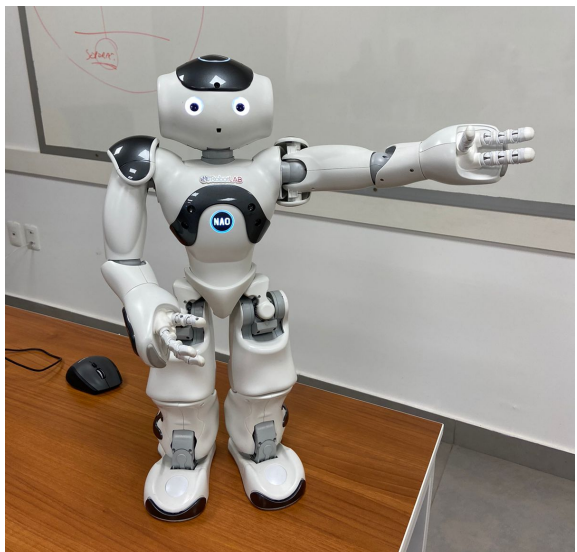

Figura 1. Robot NAO V6.

Su presentación amigable en cuestión de formato de face y cuerpo, demuestra cómo ha seguido la línea de producción de robots en la actualidad. La busca por integración de la robótica en servicios y cotidiano fue el principal enfoque de empresas de desarrollo en los últimos años, para esto, estos robots están diseñados específicamente para interactuar con los seres humanos de una manera que sea cómoda, segura y agradable. Como esta área está en constante evolución, es importante considerar tanto las interesantes oportunidades como los desafíos éticos y prácticos que pueden surgir.

## 2.2 Programación de la simulación

Para la implementación de la idea de simulación se utilizó una plataforma llamada Node Red para programar el robot NAO. Esta herramienta permite configurar las interacciones específicas que realizará el robot con los humanos. Node Red actúa como un entorno de programación visual que facilita la creación y personalización de acciones del robot. Ella permite conectar dispositivos, servicios y API (Interfaz de programación de aplicaciones) de forma intuitiva y eficiente. Node Red está construido en Node.js y utiliza un enfoque de programación basado en procesos, lo que significa que puede crear aplicaciones como diagramas de flujo visuales en lugar de escribir código tradicional mientras permite fácil integración con potentes lenguajes de programación como Python V3. Además, con el avance de la tecnología, ya posibilita una conexión con GPT, una inteligencia artificial para el procesamiento del lenguaje natural.

La integración de GPT puede enriquecer la interacción del robot NAO con los visitantes como forma de ampliar las ramas de aplicaciones de esta tecnología. Utilizando GPT, por ejemplo, el robot puede proporcionar respuestas y explicaciones más detalladas sobre las obras de arte en el museo simulado, pudiendo responder de manera más actualizada acerca de descubrimientos históricos. Esto agrega una capa adicional de información y contexto a los visitantes, lo que genera una experiencia más rica y educativa, en la Figura 2 en continuación es posible visualizar una imagen de la utilización de la herramienta de programación Node-RED conectado a el robot NAO.

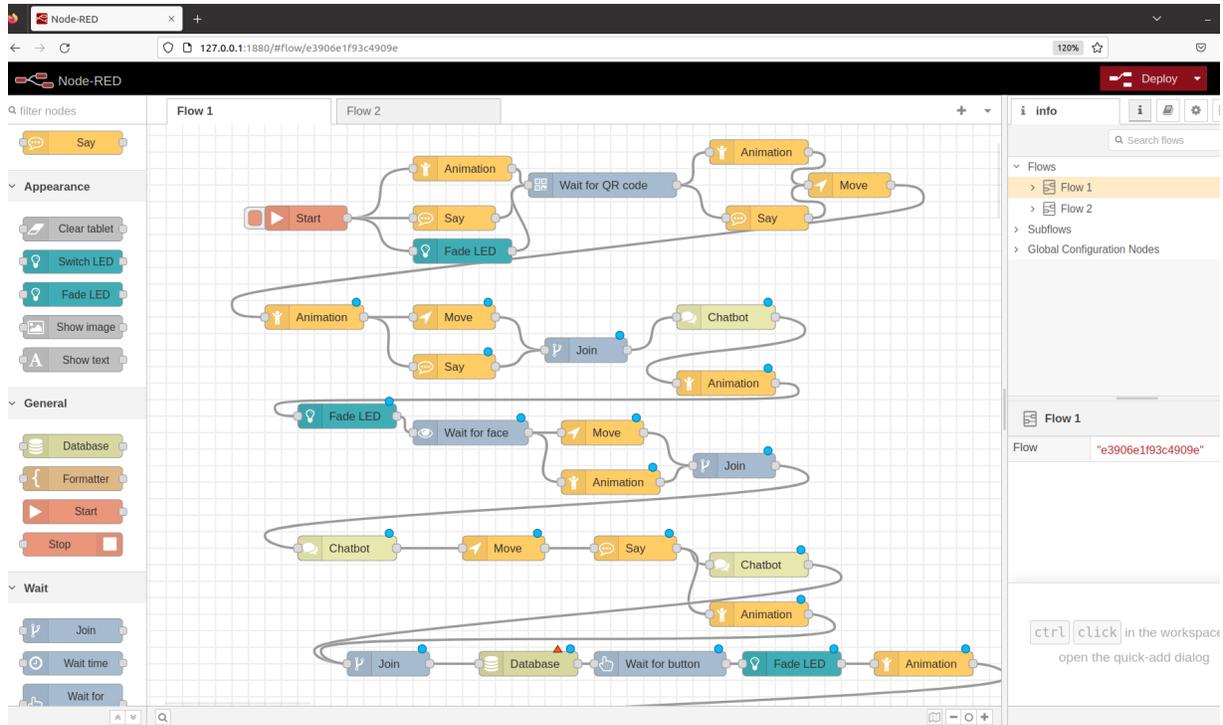

Figura 2. Pantalla de utilización de la herramienta Node-Red.

## 2.3 - Ambiente Simulado

Como se ilustra en la Figura 1, Node-RED se basa en una interfaz web que permite arrastrar y soltar nodos para crear flujos donde cada nodo tiene una funcionalidad específica. Por ejemplo, puede existir un nodo para ejecutar ciertos scripts de Python, recolectar información del entorno mediante sensores, procesarla a través de modelos de IA e interactuar con el entorno.

La conexión de los nodos en un flujo determina la operación del robot. De esta manera, se utilizó esta herramienta para simular un espacio que representa la disposición básica de obras de arte en un museo. Se dispusieron reproducciones a escala reducida de pinturas y esculturas siguiendo un recorrido preestablecido.Una imagen de esta simulación se puede visualizar en la Figura 3.

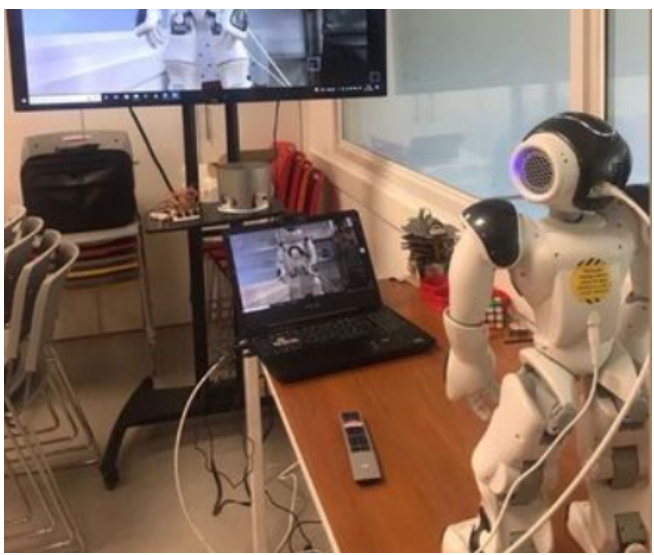

Figura 3. Ambiente Simulado.

**2.4 - Configuración del Robot NAO y Chatbot:**

El robot NAO fue programado para asumir el rol de guía, utilizando movimientos y gestos para interactuar con los estudiantes participantes. Se diseñaron respuestas verbales para responder a las preguntas comunes, utilizando en este caso, el recurso del chatbot para proporcionar información detallada sobre las obras de arte en el entorno simulado. Durante de algunas de las simulaciones de movimientos por el museo el robot NAO puede explicar a los visitantes acerca de dónde están, una imagen de esta simulación se ve en la Figura 3, la idea de utilización del chatbot posibilita el uso de inteligencia artificial de manera integrada con la herramienta Node-Red, donde un bloque específico ya contiene adentro todos los procesos necesarios para la conexión con la IA.

Se sabe que la integración de robots con inteligencia artificial se ha convertido en un tema de amplia discusión gracias al avance de herramientas como Chat GPT y otras que facilitan una interacción ágil entre el usuario y sus necesidades. Estas herramientas de Inteligencia Artificial posibilitan un primer contacto con la tecnología, mostrando un enorme potencial cuando se aplican a robots, cuyas interacciones se tornan cada vez más orgánicas. El ejemplo presentado por este proyecto muestra cómo se puede utilizar la funcionalidad de forma integrada a los robots, facilitando así, un trabajo complejo de en este caso tener una gran base de datos acerca del acervo de obras del museo entero.

Como la inteligencia artificial está enteramente integrada a la internet y todo los datos que circulan entre ella, existen entonces, un cierto cuidado que hay que tener perante a la veracidad de las informaciones que se van a extraer via herramienta ChatBot. Por este motivo, la utilización de esta herramienta para la situación propuesta necesita de aprimoramiento para el filtraje de informaciones y también acerca de la validad de esa información durante la época de las obras y nuevos descubrimientos acerca de ellas.

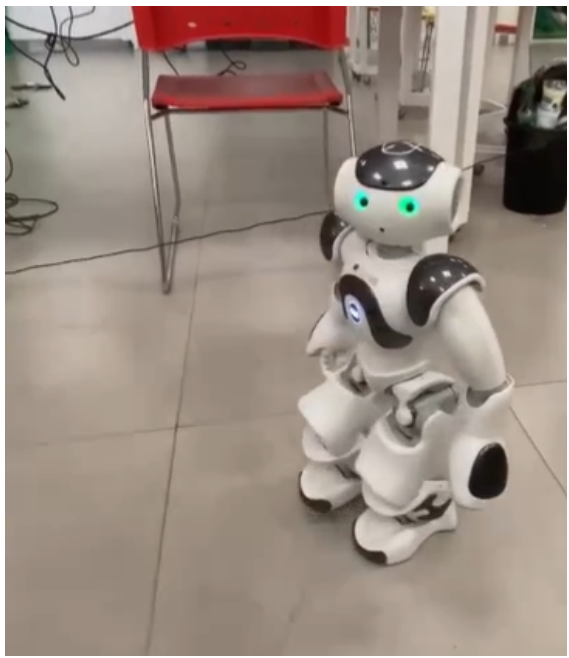

Figura 3. Pruebas del escenario con el Robot NAO v6.

**2.5 - Participantes Colaboradores**

Como la idea de este proyecto de simulación de interacción humano-robot partió de una secuencia de talleres con el Docente  Tobias Doernbach de la Universidad de Ostfalia en Alemania en el ITR Norte de Rivera, un selecto grupo de siete estudiantes y tres docentes participaron en la investigación, contribuyendo en el ensayo de simulación. El ejercicio dio a los estudiantes la oportunidad de interactuar con el robot NAO y también pudieron interactuar con el chatbot. Cada equipo tuvo la disponibilidad de elegir un escenario para simulación, teniendo en cuenta necesidades de la sociedad en cuestión de automatización de servicios, partiendo de esto surgió la idea de investigación acerca de simulación en un museo. Una imagen de estas simulaciones se encuentra en la Figura 4 a continuación.

Durante la simulación, los estudiantes colaboradores interactuaron de manera auténtica con el robot NAO y el chatbot en el escenario simulado del museo. Sus preguntas, reacciones y observaciones proporcionaron una valiosa perspectiva sobre la efectividad y la calidad de la interacción. La participación de los estudiantes no solo contribuyó a la recopilación de datos, sino que también fomentó un ambiente de cooperación y aprendizaje conjunto entre los colaboradores y el equipo de investigación. Un trabajo como este enriquece la expansión del conocimiento académico en el área de robótica, con la experiencia de un docente renombrado en el seguimiento, fue posible aprimorar estas habilidades de integrar la robótica al mundo cotidiano.

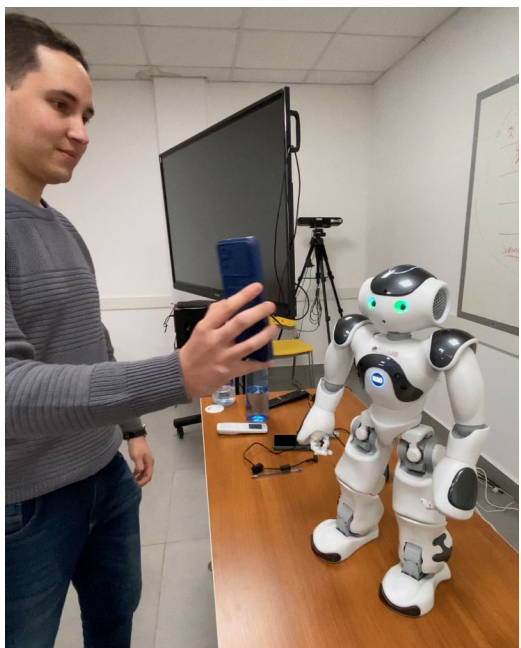

Figura 4 - Interacción de colaboradores con el robot NAO.

## 3 - RESULTADOS Y DISCUSIÓN

Durante la simulación de la aplicación de robots humanoides como guías en un museo de arte, se recopilaron datos que arrojaron luz sobre la efectividad de esta tecnología en la interacción con los visitantes. A continuación, se presentan los hallazgos clave derivados de la simulación y se discuten en el contexto de las ventajas y desafíos asociados con esta tecnología.

### 3.1 - Efectividad de la Información Proporcionada

Tanto el robot NAO como el chatbot demostraron ser efectivos para proporcionar información básica sobre las obras de arte presentes en el museo simulado. Los estudiantes colaboradores expresaron satisfacción con la información proporcionada, destacando la utilidad de las respuestas tanto visuales como verbales, como se puede ver en la Figura 5 la reacción de los visitantes.

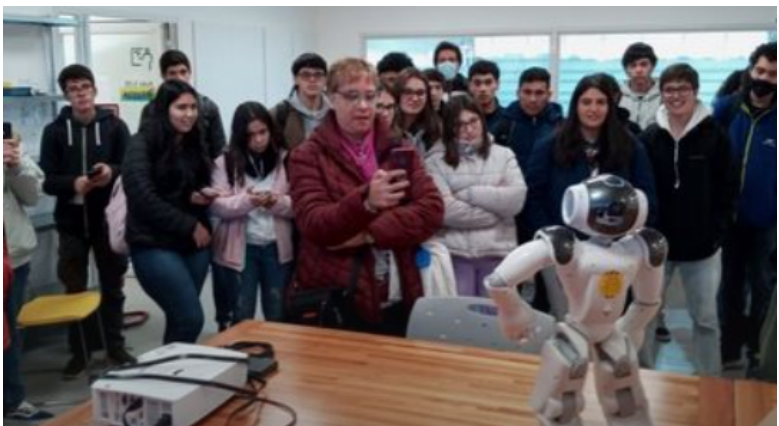

Figura 5 - Interacción de visitantes con el robot NAO V6.

### 3.2 - Interacción Natural

Las interacciones entre los estudiantes colaboradores y el robot NAO fueron en gran medida fluidas y naturales. Los gestos y movimientos programados del robot contribuyeron a una experiencia auténtica y atractiva (como se ve en la Figura 6, el robot haciendo un gesto), facilitando la orientación y el aprendizaje.

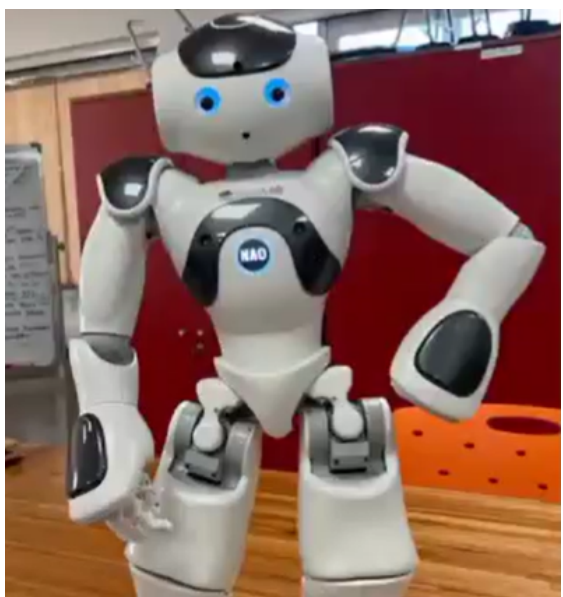

Figura 6 - Robot NAO V6 haciendo un gesto.

A pesar de la interacción positiva, se identificaron áreas de mejora. Algunos estudiantes colaboradores notaron ocasionalmente respuestas inexactas o ambiguas del chatbot. Esto resalta la importancia de continuar perfeccionando la precisión y calidad de la información proporcionada.

Las respuestas imprecisas o ambiguas del chatbot podrían afectar la confianza de los visitantes en la información brindada por el robot NAO. La exactitud y coherencia de la información son esenciales para una experiencia educativa exitosa. Abordar este desafío podría requerir revisar y mejorar constantemente la base de datos y los algoritmos del chatbot, así como establecer métodos para corregir errores rápidamente.

Este estudio destaca la importancia de ver el proceso de desarrollo como continuo y en evolución. La retroalimentación de los usuarios, como los estudiantes colaboradores aquí, es invaluable para identificar problemas y áreas de mejora que podrían pasarse por alto en el desarrollo inicial. A través de ajustes constantes en la programación y configuración, es posible lograr un mayor nivel de precisión y confiabilidad en las respuestas del chatbot, mejorando la calidad general de la interacción.

### 3.4 - Verificación de Bugs y Mejora de la Interacción

Fue crucial para esta investigación identificar y corregir los posibles problemas técnicos así como los errores en la interacción entre el robot NAO, el chatbot y los estudiantes colaboradores. Con el fin de garantizar un funcionamiento fluido, se ha realizado una exhaustiva revisión. Con base en los comentarios iniciales, se efectuaron modificaciones en la programación con el fin de perfeccionar la naturaleza y calidad de las interacciones llevadas a cabo con los visitantes simulados.

La simulación sugiere que la presencia de robots humanoides puede enriquecer la experiencia del visitante al ofrecer una nueva forma de explorar y aprender sobre las obras de arte. Sin embargo, también subraya la importancia de mantener un equilibrio entre la tecnología y la interacción humana. En la Figura 7 se puede ver al robot NAO V6 hablando.

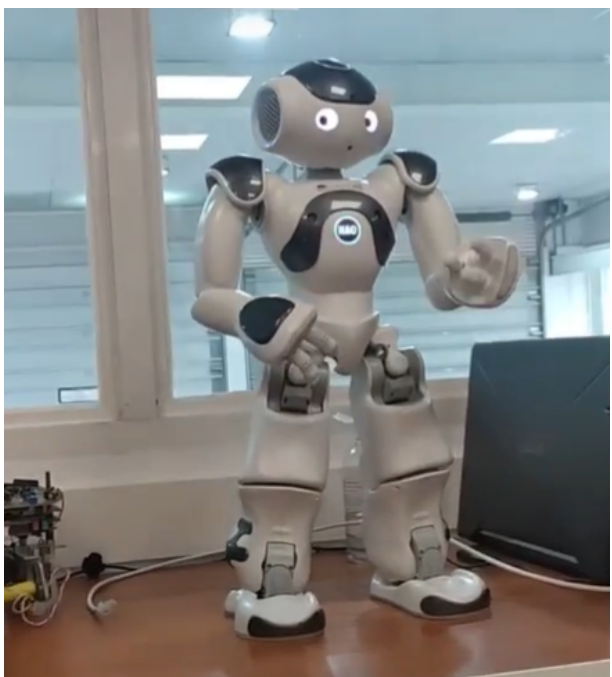

Figura 7 - Robot NAO V6 hablando.

## 4 - CONCLUSIONES

La simulación realizada, que evaluó el uso de robots humanoides como guías interactivas en un museo de arte mediante la programación del robot NAO y la integración de un chatbot, aportó información valiosa sobre el potencial de esta tecnología en contextos culturales. Los resultados de la simulación mostraron una participación de los estudiantes colaboradores en las interacciones con el robot NAO y el chatbot. La eficacia a la hora de transmitir información básica sobre las obras de arte combinada con una interacción gestual natural entre los estudiantes y el robot sugiere que la presencia de robots humanoides puede enriquecer la experiencia del visitante en un museo de arte.

Sin embargo, también destacó desafíos a abordar, como la necesidad de mejorar la precisión de las respuestas proporcionadas por el chatbot. Estos desafíos resaltan la importancia de la mejora continua en la programación y configuración de robots para lograr una interacción de alta calidad.

En última instancia, este estudio contribuye a comprender las oportunidades y desafíos en la implementación de tecnologías humanoides en contextos culturales. La simulación sirve como base para futuras investigaciones destinadas a mejorar la interacción entre humanos y robots en entornos de museos. La combinación de Node Red y Chat GPT en este estudio ilustra el potencial de integrar múltiples tecnologías para mejorar las experiencias de los visitantes y fomentar un mayor compromiso con las artes y la cultura.

## 5- REFERENCIAS

Hellou, Mehdi et al. Technical Methods for Social Robots in Museum Settings: An Overview of Literature. International Journal of Social Robotics, v. 14, n. 8, p. 1767-1786, 2022.

Gasteiger N, Hellou M, Ahn HS (2021) Deploying social robots in museum settings: a quasi-systematic review exploring purpose and acceptability. Int J Adv Rob Syst 18(6):17298814211066740

Irfan B et al (2020) Challenges of a real-world HRI study with non-native English speakers: can personalisation save the day? pp 272–274

Bansal, H., Khan, R.: A review paper on human computer interaction. Int. J. Adv. Res. Comput. Sci. Softw. Eng. 8, 53 (2018). https://doi.org/10.23956/ijarcsse.v8i4.630

TLILI, Ahmed et al. What if the devil is my guardian angel: ChatGPT as a case study of using chatbots in education. Smart Learning Environments, v. 10, n. 1, p. 15, 2023.